\RequirePackage{iftex}
\RequirePDFTeX
\documentclass[letterpaper,10pt,conference]{ieeeconf}
\IEEEoverridecommandlockouts
\usepackage{graphicx}
\usepackage{capt-of}
\usepackage{booktabs}
\usepackage{float}
\usepackage{needspace}
\usepackage{tabularx}
\usepackage{multirow}
\usepackage{cite}
\usepackage{amsmath,amssymb,amsfonts,mathtools}
\usepackage{algorithmic}
\usepackage{textcomp}

\usepackage{amsthm}

\theoremstyle{plain}

\theoremstyle{definition}

\theoremstyle{remark}

\usepackage{dsfont}
\usepackage{xcolor}
\usepackage{colortbl}
\usepackage{etoolbox}

\makeatletter
\patchcmd{\@makecaption}{\scshape}{\upshape}{}{%
    \PackageError{SPAR}{Could not update table caption font}{Check the ieeeconf caption definition.}}
\patchcmd{\@makecaption}{\begin{center}}{\vspace*{4bp}\begin{center}}{}{%
    \PackageError{SPAR}{Could not adjust table caption clearance}{Check the ieeeconf caption definition.}}
\makeatother

\definecolor{darkgreen}{RGB}{0,150,0}

\usepackage[hidelinks]{hyperref}
\usepackage[capitalize,noabbrev]{cleveref}
\crefname{figure}{Fig.}{Figs.}
\Crefname{figure}{Fig.}{Figs.}

\begin{document}
\bstctlcite{IEEEreferencecontrol}

\title{\LARGE\bfseries SPAR: Structural Pruning and Action Recovery for Vision-Language-Action Models}

\author{
    \authorblockN{Shenli Zheng$^{1}$, Zhou Fang$^{1}$, Yuan Du$^{1}$, Yijiang Liu$^{2}$, Li Du$^{1,*}$}
    \authorblockA{$^{1}$Nanjing University, Nanjing, China\\
    $^{2}$Nanjing University of Information Science and Technology, Nanjing, China\\
    \texttt{shenlizheng@smail.nju.edu.cn}, \texttt{zfang@nju.edu.cn}\\
    \texttt{yuandu@nju.edu.cn}, \texttt{liuyijiang@nuist.edu.cn}\\
    $^{*}$Corresponding author: \texttt{ldu@nju.edu.cn}}
}
\hypersetup{
    pdftitle={SPAR: Structural Pruning and Action Recovery for Vision-Language-Action Models},
    pdfauthor={Shenli Zheng, Zhou Fang, Yuan Du, Yijiang Liu, Li Du},
    pdfsubject={arXiv preprint},
    pdfkeywords={vision-language-action models, structural pruning, reinforcement learning}
}
\IEEEaftertitletext{\begin{center}
\begin{minipage}{\textwidth}
    \centering
    \includegraphics[width=\linewidth]{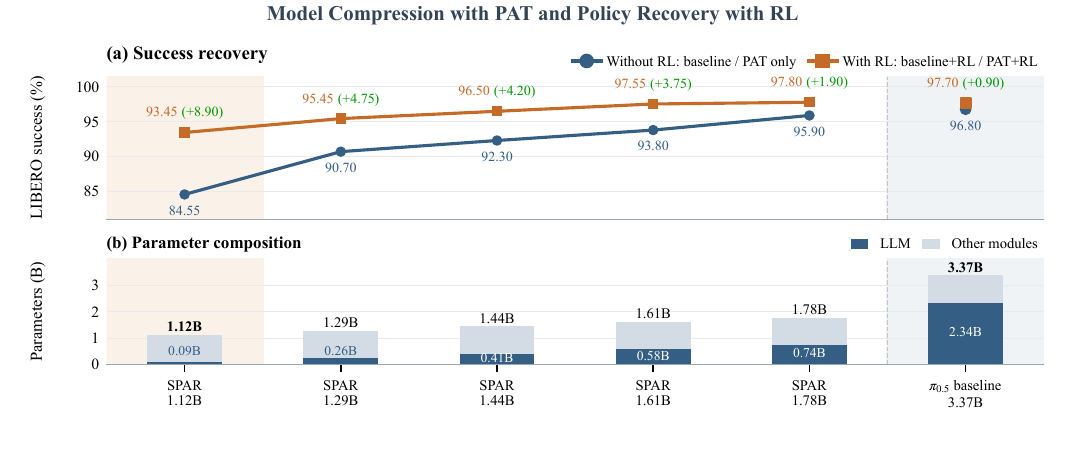}
    \captionof{figure}{\textbf{Model Compression and Policy Recovery on LIBERO.} (a) Average success rates before and after RL, with gains shown in green (percentage points). (b) Parameter counts of the pruned models and the unpruned baseline (right), broken down into the LLM backbone and remaining components. With 0.09B LLM parameters (1.12B total), SPAR reaches 93.45\% success after RL, an 8.90-percentage-point gain over its PAT-only counterpart.}
    \label{fig:accuracy_comparison}
\end{minipage}
\end{center}
}
\maketitle
\pagestyle{plain}
\thispagestyle{plain}
\begin{figure*}[!t]
    \centering
    \includegraphics[width=\linewidth]{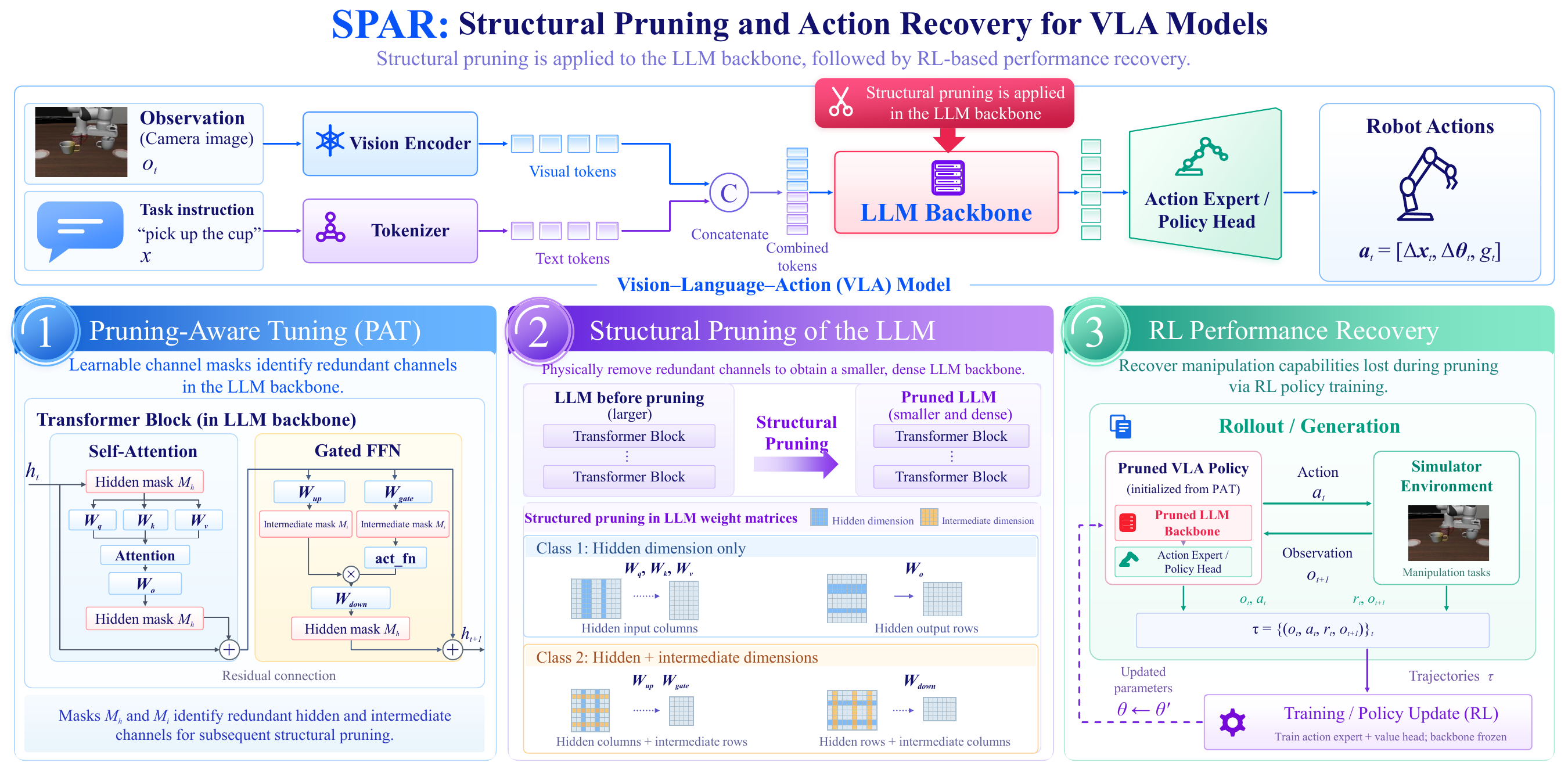}
    \caption{\textbf{Overview of the SPAR Framework.}
    \textbf{Top:} SPAR compresses the LLM backbone of a VLA policy, which integrates visual and language inputs to guide the action expert.
    \textbf{Bottom:} \textbf{(1)} Through \textbf{pruning-aware tuning}, SPAR jointly learns channel masks for the hidden dimension and the FFN intermediate dimension while adapting the policy to manipulation tasks. \textbf{(2)} These masks guide \textbf{structural pruning}, which removes inactive channels to form a smaller dense backbone. \textbf{(3)} After supervised fine-tuning of the compact policy, \textbf{RL recovery} further improves task performance through simulated interaction, updating the action expert and value head while keeping the pruned LLM backbone frozen.}
    \label{fig:pipeline}
\end{figure*}

\begin{abstract}
The limited onboard computing resources of robots make the inference demands of vision-language-action (VLA) models costly to accommodate. Reducing these demands through aggressive compression, however, can degrade manipulation performance beyond what supervised fine-tuning alone recovers. We present SPAR (Structural Pruning and Action Recovery), which simultaneously prunes the hidden and feed-forward intermediate dimensions of the large language model (LLM) backbone through pruning-aware tuning, then applies reinforcement learning (RL) to recover policy performance. Applied to $\pi_{0.5}$, SPAR reduces the language backbone from 2.34B to 0.09B parameters and the full model from 3.37B to 1.12B, removing approximately 96\% and 67\% of their parameters, respectively. At this compression level, RL raises average success across the four LIBERO suites from 84.55\% to 93.45\%, an 8.90-percentage-point gain. The compressed policy achieves a $\boldsymbol{1.98\times}$ full-policy inference speedup over the unpruned model. These results identify substantial intermediate-dimension redundancy and show that post-pruning RL can recover manipulation performance in compact policies.
\end{abstract}

\section{Introduction}
Vision-language-action (VLA) models such as RT-2~\cite{brohan2023rt2visionlanguageactionmodelstransfer}, OpenVLA~\cite{kim2024openvlaopensourcevisionlanguageactionmodel}, and $\pi_{0.5}$~\cite{intelligence2025pi05visionlanguageactionmodelopenworld} transfer pretrained representations to robotic manipulation, but their inference demands strain limited onboard compute. In $\pi_{0.5}$, the LLM accounts for 2.34B of 3.37B parameters. This motivates a central question: \textit{how much language-model capacity do manipulation tasks require, and how can a smaller policy retain useful capabilities?}

Efficiency approaches include compact policies such as SmolVLA~\cite{shukor2025smolvlavisionlanguageactionmodelaffordable} and NORA~\cite{hung2025norasmallopensourcedgeneralist}, preference-based adaptation in NORA-1.5~\cite{hung2025nora15}, and synthetic pretraining in GraspVLA~\cite{deng2025graspvla}. For existing policies, SP-VLA~\cite{li2025spvlajointmodelscheduling} and DART~\cite{wen2025stoplookingimportanttokens} reduce tokens, while EfficientVLA~\cite{yang2025efficientvlatrainingfreeaccelerationcompression} and MoLe-VLA~\cite{zhang2025molevladynamiclayerskippingvision} reduce layer computation. Token reduction alone leaves backbone weights intact. EcoVLA~\cite{huang2026ecovla} dynamically selects channels and attention heads; we instead permanently reduce backbone dimensions to produce a smaller dense model.

Permanent compression introduces a second challenge: recovering manipulation capabilities lost when parameters are removed. RLRC~\cite{chen2026rlrc} combines structured pruning with supervised fine-tuning (SFT) and reinforcement learning (RL), while SimpleVLA-RL~\cite{li2025simplevlarlscalingvlatraining} demonstrates that interaction can improve policies beyond SFT. These results motivate studying how the allocation of a small parameter budget and adaptation during pruning affect subsequent recovery. The hidden dimension couples representations across Transformer blocks, whereas the intermediate dimension controls expansion within feed-forward layers. Their different roles motivate separate retention targets for the two dimensions.

We introduce \textbf{SPAR} (Structural Pruning and Action Recovery), which couples dimension-aware pruning with post-pruning RL to recover task performance (\Cref{fig:pipeline}). Building on pruning-aware tuning (PAT)~\cite{liu2025pat}, SPAR jointly learns masks for the hidden and intermediate dimensions with separate retention targets and a regularized mask-learning objective. Channel selection and supervised adaptation are performed together, followed by physical pruning and fine-tuning of the reduced model. The resulting compact policy provides the initialization for RL, which further improves manipulation performance through interaction.

Our evaluation on LIBERO~\cite{NEURIPS2023_8c3c6668} and LIBERO-Plus~\cite{fei2025liberoplus} examines both task success and robustness under compression. Pruning only the intermediate dimension preserves high task success, while additional hidden-dimension pruning at a fixed intermediate-dimension retention target causes larger losses that RL can substantially mitigate. As a representative result, SPAR achieves 93.45\% average LIBERO success with a 0.09B language backbone (1.12B total), approximately 4\% of the original LLM size (\Cref{fig:accuracy_comparison}). These findings support dimension-aware compression and recovery as complementary tools for compact robotic policies.

Our main contributions are:
\begin{itemize}
\item A \textbf{dual-dimension pruning framework} that jointly learns masks for the hidden and intermediate dimensions with separate retention targets and a regularized mask-learning objective.
\item An \textbf{empirical analysis of dimension sensitivity}: pruning only the intermediate dimension preserves high task success, while additional hidden-dimension pruning causes larger performance losses in our evaluated tasks.
\item An \textbf{evaluation of recovery and efficiency} across compression levels, showing improvements in task success and robustness after RL, together with measured full-policy inference acceleration.
\end{itemize}

\section{Related Work}
\subsection{Vision-Language-Action (VLA) models}
VLA models connect pretrained vision-language representations to robotic control. OpenVLA~\cite{kim2024openvlaopensourcevisionlanguageactionmodel} predicts discretized actions autoregressively, whereas OpenVLA-OFT~\cite{kim2025finetuningvisionlanguageactionmodelsoptimizing} uses parallel decoding and action chunking. CogACT~\cite{li2024cogactfoundationalvisionlanguageactionmodel} separates action generation into a diffusion expert. The $\pi_0$ family~\cite{black2025pi0visionlanguageactionflowmodel} couples a PaliGemma backbone~\cite{beyer2024paligemmaversatile3bvlm} with a flow-matching action expert; $\pi_{0.5}$~\cite{intelligence2025pi05visionlanguageactionmodelopenworld} adds semantic subtask prediction. TinyVLA~\cite{wen2025tinyvla} instead combines a compact multimodal backbone with a diffusion decoder.

\subsection{Pruning for VLA models}
LLM compression includes weight sparsification in SparseGPT~\cite{frantar2023sparsegpt}, coupled-structure removal in LLM-Pruner~\cite{ma2023llmpruner}, and row/column deletion to form smaller dense matrices in SliceGPT~\cite{ashkboos2024slicegpt}. PAT~\cite{liu2025pat} jointly performs hidden-dimension pruning and fine-tuning through trainable channel masks and Hybrid Sparsification Modules. For VLA models, DART~\cite{wen2025stoplookingimportanttokens} and SP-VLA~\cite{li2025spvlajointmodelscheduling} reduce visual tokens; EfficientVLA~\cite{yang2025efficientvlatrainingfreeaccelerationcompression} removes low-impact layers, and MoLe-VLA~\cite{zhang2025molevladynamiclayerskippingvision} dynamically skips layers. EcoVLA~\cite{huang2026ecovla} adaptively selects channels and attention heads with parallel sparsity updates and custom kernels. Building on PAT, SPAR learns masks for both the hidden and intermediate dimensions with separate retention targets, physically reduces the dense backbone, and applies RL recovery. Our evaluation examines the effects of compression on both manipulation success and robustness.

\subsection{Reinforcement learning for VLA models}
RL for VLA models includes both offline and online approaches. Offline RL learns from a fixed dataset without additional environment interaction, whereas online RL collects new experience as training proceeds. The RECAP recipe for $\pi_{0.6}^*$~\cite{intelligence2025pi06vlalearnsexperience} starts with offline RL pretraining, then alternates robot data collection with offline updates using advantage-conditioned policies. For online VLA training, SimpleVLA-RL~\cite{li2025simplevlarlscalingvlatraining} optimizes policies through environment interaction with outcome-based rewards, while RLinf~\cite{zang2025rlinfvlaunifiedefficientframework} provides scalable training infrastructure. In the compression setting, RLRC~\cite{chen2026rlrc} combines structured pruning, supervised fine-tuning, and RL to recover policy performance.

\subsection{Manipulation Benchmarks}
LIBERO~\cite{NEURIPS2023_8c3c6668} evaluates knowledge transfer across language-conditioned manipulation tasks. ManiSkill2~\cite{gu2023maniskill2} offers 20 task families spanning stationary and mobile manipulation, single- and dual-arm robots, and rigid and soft objects. Robustness-oriented extensions probe behavior beyond standard task completion: LIBERO-PRO~\cite{zhou2025liberopro} perturbs manipulated objects, initial states, instructions, and environments to examine sensitivity to memorized task conditions. LIBERO-Plus~\cite{fei2025liberoplus} evaluates seven perturbation dimensions, including camera viewpoints, robot initial states, and scene appearance. We use LIBERO to evaluate task success and LIBERO-Plus to assess robustness under perturbations after compression.

\section{Methodology}

SPAR combines structural pruning with post-pruning RL (\Cref{fig:pipeline}). We adapt pruning-aware tuning (PAT)~\cite{liu2025pat} to jointly learn hidden and intermediate channel masks with separate retention targets. After mask learning, we physically remove inactive channels and finetune the compact policy before RL recovery. In our experiments, \emph{PAT only} denotes this pruning and supervised adaptation stage within SPAR, and \emph{PAT+RL} denotes the complete framework.

\paragraph{PAT background}
PAT's Hybrid Sparsification Module~\cite{liu2025pat} combines channel selection with adaptation of pretrained weights. For a linear map $\mathbf{W}\in\mathbb{R}^{d_o\times d_i}$, its forward operation can be written as
\begin{equation}
    \mathbf{Z}=\operatorname{diag}(\mathbf{M})\mathbf{D}\mathbf{W}\mathbf{X},
\end{equation}
where $\mathbf{M}$ selects output channels and $\mathbf{D}$ is the knowledge-transformation operator. The original formulation regularizes channel sparsity and the low-rank transformation's orthogonality while optimizing the supervised objective. This joint optimization allows pretrained representations to adapt as channel selection changes. SPAR adopts the pruning-aware training principle and introduces separate retention targets for the hidden and intermediate dimensions, a sigmoid mask schedule, and a differentiable count penalty. The following subsections formulate these channel-selection components and their role in preparing the compact policy for recovery.

\subsection{Dual-Dimension Structural Pruning}

The hidden dimension determines the width of the representation passed between Transformer blocks, whereas the intermediate dimension controls the expansion within each feed-forward network (FFN). Pruning the hidden dimension therefore requires coordinated changes to the attention and FFN projections to preserve residual connections. In contrast, pruning the intermediate dimension reduces the FFN expansion while leaving the residual-stream width unchanged. SPAR uses separate retention targets for the two dimensions to control these distinct aspects of model capacity.

Let $d_{model}$ and $d_{ffn}$ denote the hidden dimension and the FFN intermediate dimension, respectively. We learn a mask for each dimension, $\mathbf{M}_{hid}\in[0,1]^{d_{model}}$ and $\mathbf{M}_{int}\in[0,1]^{d_{ffn}}$, with retention targets $\rho_{hid}$ and $\rho_{int}$. Both masks are shared across layers. Writing their diagonal forms as $\mathbf{D}_h=\operatorname{diag}(\mathbf{M}_{hid})$ and $\mathbf{D}_i=\operatorname{diag}(\mathbf{M}_{int})$, we apply the hidden-dimension mask to the attention projections:
\begin{equation}
\begin{aligned}
    \mathbf{W}'_{\{Q,K,V\}} &= \mathbf{W}_{\{Q,K,V\}}\mathbf{D}_h, \\
    \mathbf{W}'_O &= \mathbf{D}_h\mathbf{W}_O.
\end{aligned}
\end{equation}
The FFN projections depend on both dimensions and are masked as
\begin{equation}
\begin{aligned}
    \mathbf{W}'_{\{up,gate\}} &= \mathbf{D}_i\mathbf{W}_{\{up,gate\}}\mathbf{D}_h, \\
    \mathbf{W}'_{down} &= \mathbf{D}_h\mathbf{W}_{down}\mathbf{D}_i.
\end{aligned}
\end{equation}
In $\mathbf{W}_{up}$ and $\mathbf{W}_{gate}$, rows correspond to the intermediate dimension and columns to the hidden dimension; this arrangement is reversed in $\mathbf{W}_{down}$. Sharing the hidden-dimension mask across blocks ensures that their residual connections use the same retained channels. Once the masks are binarized, we remove the rows and columns associated with inactive channels. The resulting backbone uses smaller dense weight matrices while preserving the visual-token sequence and Transformer depth.

\subsection{Mask Learning and Regularization}

At low retention, mask learning must select a small set of channels while preserving task-relevant representations. We combine continuous gates for gradient-based adaptation, a count penalty for the retention target, and binary regularizers for decisive channel selection. This formulation allows the supervised objective to guide channel importance while progressively preparing the masks for physical pruning.

\paragraph{Soft-to-hard masking}
For either mask, trainable mask logits $\mathbf{W}_M$ are initialized to $+1.5$ for an initial candidate set of retained channels and $-1.5$ for the remaining channels. These assignments remain learnable during supervised adaptation. During training,
\begin{equation}
    \mathbf{M}(s)=\sigma\!\left(\tau(s)\mathbf{W}_M\right),
\end{equation}
where $\sigma$ is the logistic sigmoid and $s$ is the training step. The inverse temperature increases linearly from 1 to 10:
\begin{equation}
    \tau(s)=1+9\min\!\left(\frac{s}{r_{stop}s_0},1\right),
\end{equation}
where $s_0$ is the reference training horizon for the schedule and $r_{stop}$ controls its endpoint (typically 0.8). The inverse temperature reaches 10 at step $r_{stop}s_0$, progressively sharpening the mask toward binary values. After mask learning, we set $\mathbf{M}_{bin}=\mathds{1}[\mathbf{W}_M\geq0]$ and use the resulting channel selection for physical pruning.

The gate's derivative makes the optimization trade-off explicit:
\begin{equation}
    \frac{\partial M_j}{\partial W_{M,j}}=\tau(s)M_j(1-M_j).
\end{equation}
The supervised loss updates channel selection through these gradients, with larger responses near the decision boundary and smaller responses near the binary endpoints. Annealing progressively sharpens the masks as the policy adapts. Once the masks are thresholded, the retained channel set is fixed and supervised fine-tuning adapts the reduced dense matrices to the manipulation tasks.

\begin{table*}[!t]
    \centering
    \caption{\textbf{Effect of FFN Intermediate-Dimension Pruning on LIBERO Performance.}
    We vary the FFN intermediate-dimension retention target in $\pi_{0.5}$ while keeping the hidden dimension unchanged, and evaluate each policy after SPAR's PAT stage without RL. The ratios $\rho_{hid}/\rho_{int}$ specify retention targets for the hidden and intermediate dimensions. Parameter counts are in billions, and pruning ratios indicate the approximate percentages of parameters removed from the full model and LLM relative to the unpruned baseline. Success rates are reported as percentages, with Avg. denoting the mean across the four suites and bold highlighting the best score in each column.}
    \label{tab:main_results_intermediate}
    \footnotesize
    \setlength{\tabcolsep}{2pt}
    \begin{tabular*}{\textwidth}{@{\extracolsep{\fill}}llcccccccccc@{}}
        \toprule
        \multirow{2}{*}{\textbf{Model}} & \multirow{2}{*}{\textbf{Method}} & \multirow{2}{*}{$\boldsymbol{\rho_{hid}/\rho_{int}}$} & \multirow{2}{*}{\shortstack{\textbf{Params (B)}\\\textbf{Total / LLM}}} & \multirow{2}{*}{\textbf{Speedup ($\uparrow$)}} & \multicolumn{2}{c}{\textbf{Pruning Ratio}} & \multirow{2}{*}{\textbf{Spatial}} & \multirow{2}{*}{\textbf{Object}} & \multirow{2}{*}{\textbf{Goal}} & \multirow{2}{*}{\textbf{Long}} & \multirow{2}{*}{\textbf{Avg.}} \\
        \cmidrule(lr){6-7}
        & & & & & \textbf{Model} & \textbf{LLM} & & & & & \\
        \midrule
        $\pi_{0.5}$ & baseline & 100\% / 100\% & 3.37 / 2.34 & 1.00 & 0\% & 0\% & 98.80 & \textbf{98.80} & 97.40 & 92.20 & 96.80 \\
        \midrule
        \multirow{6}{*}{$\pi_{0.5}$} & \multirow{6}{*}{\textbf{PAT only}} & 100\% / 90\% & 3.20 / 2.17 & 1.05 & 5\% & 7\% & \textbf{99.80} & 98.20 & 96.60 & 94.60 & 97.30 \\
        & & 100\% / 70\% & 2.86 / 1.83 & 1.17 & 15\% & 22\% & 98.40 & 98.00 & 98.00 & \textbf{96.20} & \textbf{97.65} \\
        & & 100\% / 50\% & 2.52 / 1.49 & 1.33 & 25\% & 36\% & 98.40 & 98.60 & \textbf{98.20} & 94.40 & 97.40 \\
        & & 100\% / 30\% & 2.19 / 1.16 & 1.52 & 35\% & 51\% & 98.80 & 98.20 & 97.80 & 94.60 & 97.35 \\
        & & 100\% / 20\% & 2.02 / 0.99 & 1.65 & 40\% & 58\% & 97.80 & 98.00 & 97.40 & 93.40 & 96.65 \\
        & & 100\% / 10\% & 1.85 / 0.82 & 1.73 & 45\% & 65\% & 97.40 & 97.60 & 96.80 & 94.00 & 96.45 \\
        \bottomrule
    \end{tabular*}
\end{table*}

\begin{table*}[!t]
    \centering
    \caption{\textbf{LIBERO Performance of SPAR and Comparison Methods.}
    PAT only denotes pruning and supervised adaptation, and PAT+RL adds RL recovery. Notation follows \Cref{tab:main_results_intermediate}. Parentheses show RL gains in percentage points. Bold marks the best success rates within each PAT group. EcoVLA speedups are reported on RTX 3090 with custom Triton kernels~\cite{huang2026ecovla}, whereas SPAR uses TensorRT on Jetson AGX Orin. These speedups are not same-platform comparisons.}
    \label{tab:main_results_int_hid}
    \footnotesize
    \setlength{\tabcolsep}{4pt}
    \begin{tabular}{llcccccccccc}
        \toprule
        \multirow{2}{*}{\textbf{Model}} & \multirow{2}{*}{\textbf{Method}} & \multirow{2}{*}{$\boldsymbol{\rho_{hid}/\rho_{int}}$} & \multirow{2}{*}{\shortstack{\textbf{Params (B)}\\\textbf{Total / LLM}}} & \multirow{2}{*}{\textbf{Speedup ($\uparrow$)}} & \multicolumn{2}{c}{\textbf{Pruning Ratio}} & \multirow{2}{*}{\textbf{Spatial}} & \multirow{2}{*}{\textbf{Object}} & \multirow{2}{*}{\textbf{Goal}} & \multirow{2}{*}{\textbf{Long}} & \multirow{2}{*}{\textbf{Avg.} ($\boldsymbol{\Delta}$)} \\
        \cmidrule(lr){6-7}
        & & & & & \textbf{Model} & \textbf{LLM} & & & & & \\
        \midrule
        \rowcolor{gray!12}
        \multicolumn{12}{l}{\textbf{Compact VLA Models}} \\
        \addlinespace[2pt]
        NORA~\cite{hung2025norasmallopensourcedgeneralist} & & -- & 3.00 / -- & -- & -- & -- & 92.20 & 95.40 & 89.40 & 74.60 & 87.90 \\
        NORA-1.5~\cite{hung2025nora15} & & -- & 3.00 / -- & -- & -- & -- & 98.00 & 96.00 & 95.40 & 90.50 & 95.00 \\
        SmolVLA~\cite{shukor2025smolvlavisionlanguageactionmodelaffordable} & & -- & 2.25 / -- & -- & -- & -- & 93.00 & 94.00 & 91.00 & 77.00 & 88.75 \\
        GraspVLA~\cite{deng2025graspvla} & & -- & 1.80 / -- & -- & -- & -- & -- & 94.10 & 91.20 & 82.00 & 89.10 \\
        \midrule
        \rowcolor{gray!12}
        \multicolumn{12}{l}{\textbf{VLA Pruning Methods}} \\
        \addlinespace[2pt]
        \multirow{2}{*}{$\pi_{0.5}$} & EcoVLA (25\%) & -- & -- & 1.31 & -- & -- & 98.20 & 98.60 & 98.40 & 91.60 & 96.70 \\
        & EcoVLA (37.5\%) & -- & -- & 1.46 & -- & -- & 97.80 & 98.40 & 96.80 & 87.00 & 95.00 \\
        \midrule
        \rowcolor{gray!22}
        $\pi_{0.5}$ & baseline & 100\% / 100\% & 3.37 / 2.34 & 1.00 & 0\% & 0\% & 98.80 & 98.80 & 97.40 & 92.20 & 96.80 \\
        \addlinespace[2pt]
        \multirow{5}{*}{$\pi_{0.5}$} & \multirow{5}{*}{\textbf{PAT only}} & 90\% / 10\% & 1.78 / 0.74 & 1.78 & 47\% & 68\% & \textbf{97.80} & \textbf{96.00} & \textbf{97.20} & \textbf{92.60} & \textbf{95.90} \\
        & & 70\% / 10\% & 1.61 / 0.58 & 1.81 & 52\% & 75\% & 95.80 & 94.60 & 95.60 & 89.20 & 93.80 \\
        & & 50\% / 10\% & 1.44 / 0.41 & 1.89 & 57\% & 82\% & 94.00 & 95.20 & 94.00 & 86.00 & 92.30 \\
        & & 30\% / 10\% & 1.29 / 0.26 & 1.95 & 62\% & 89\% & 94.00 & 90.60 & 94.40 & 83.80 & 90.70 \\
        & & 10\% / 10\% & 1.12 / 0.09 & 1.98 & 67\% & 96\% & 94.20 & 82.80 & 86.20 & 75.00 & 84.55 \\
        \midrule
        \addlinespace[2pt]
        \rowcolor{gray!22}
        $\pi_{0.5}$ & baseline+RL & 100\% / 100\% & 3.37 / 2.34 & 1.00 & 0\% & 0\% & 99.20 & 99.60 & 99.20 & 92.80 & 97.70\,\textcolor{darkgreen}{(+0.90)} \\
        \addlinespace[2pt]
        \multirow{5}{*}{$\pi_{0.5}$} & \multirow{5}{*}{\textbf{PAT+RL}} & 90\% / 10\% & 1.78 / 0.74 & 1.78 & 47\% & 68\% & 98.40 & 99.60 & \textbf{97.40} & \textbf{95.80} & \textbf{97.80}\,\textcolor{darkgreen}{(+1.90)} \\
        & & 70\% / 10\% & 1.61 / 0.58 & 1.81 & 52\% & 75\% & \textbf{98.60} & \textbf{99.80} & 96.80 & 95.00 & 97.55\,\textcolor{darkgreen}{(+3.75)} \\
        & & 50\% / 10\% & 1.44 / 0.41 & 1.89 & 57\% & 82\% & 97.80 & 99.60 & 96.60 & 92.00 & 96.50\,\textcolor{darkgreen}{(+4.20)} \\
        & & 30\% / 10\% & 1.29 / 0.26 & 1.95 & 62\% & 89\% & 95.40 & 98.60 & 95.60 & 92.20 & 95.45\,\textcolor{darkgreen}{(+4.75)} \\
        & & 10\% / 10\% & 1.12 / 0.09 & 1.98 & 67\% & 96\% & 97.00 & 99.20 & 93.00 & 84.60 & 93.45\,\textcolor{darkgreen}{(+8.90)} \\
        \bottomrule
    \end{tabular}
\end{table*}

\paragraph{Pruning objective}
We use three regularizers to encourage target retention and binary channel selection. For a mask with $d$ entries and target active-channel count $N_{target}$, the count penalty is
\begin{equation}
    \mathcal{L}_{count}=\lambda_1\left(\frac{\sum_{j=1}^{d}M_j-N_{target}}{N_{target}}\right)^2.
\end{equation}
The soft count $\sum_j M_j$ replaces a hard active-channel count during optimization. Its derivative with respect to any mask entry is
\begin{equation}
    \frac{\partial\mathcal{L}_{count}}{\partial M_j}
    =\frac{2\lambda_1}{N_{target}^{2}}\left(\sum_{k=1}^{d}M_k-N_{target}\right).
\end{equation}
Thus, a count above the target encourages smaller mask values, and a count below it encourages larger values. The supervised loss guides the selection of individual channels, while this penalty encourages the aggregate soft count to approach the target. To promote near-binary channel assignments, we add entropy regularization:
\begin{equation}
\begin{split}
    \mathcal{L}_{entropy}=-\lambda_2\sum_{j=1}^{d}\big[&M_j\ln(M_j+\epsilon) \\
    &+(1-M_j)\ln(1-M_j+\epsilon)\big],
\end{split}
\end{equation}
The discreteness term further encourages separation from the threshold at 0.5:
\begin{equation}
    \mathcal{L}_{disc}=\lambda_3\left(0.5-\sum_{j=1}^{d}|M_j-0.5|\right),
\end{equation}
where $\lambda_1$, $\lambda_2$, and $\lambda_3$ weight the terms and $\epsilon=10^{-7}$ stabilizes the logarithms. The training objective combines these penalties with the supervised policy fine-tuning loss $\mathcal{L}_{SFT}$:
\begin{equation}
\begin{split}
    \mathcal{L}_{total}=\mathcal{L}_{SFT}&+\mathcal{L}_{count} \\
    &+\mathcal{L}_{entropy}+\mathcal{L}_{disc}.
\end{split}
\end{equation}
Together, these soft penalties encourage near-binary masks with soft counts close to the target channel budgets, while the supervised loss guides task-relevant channel selection. We then binarize the masks, physically remove inactive channels, and continue supervised fine-tuning of the pruned model. \Cref{tab:ablation_soft_mask} evaluates each regularizer when pruning only the intermediate dimension.

\begin{figure}[!ht]
    \centering
    \includegraphics[width=\columnwidth]{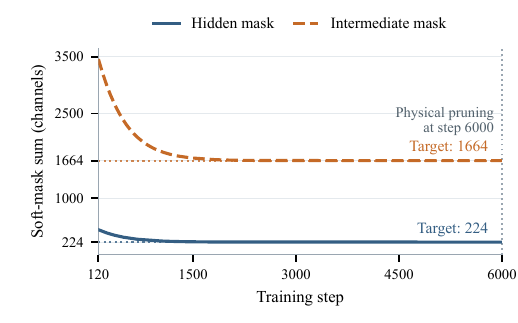}
    \caption{Soft-mask budget convergence. Hidden and intermediate soft counts approach their respective 32-channel-aligned targets of 224 and 1664 (horizontal dotted lines). The vertical line marks physical pruning at step 6000.}
    \label{fig:mask_convergence}
\end{figure}

\paragraph{Mask-budget convergence}
\Cref{fig:mask_convergence} shows $\pi_{0.5}$ mask learning on LIBERO at 10\% hidden and intermediate retention targets. Both soft counts are within 0.05\% of their aligned targets by step 3000.

\subsection{Post-Pruning RL Recovery}

After mask learning, physical pruning, and supervised fine-tuning, the compact policy may still exhibit reduced task success. We therefore use RL to improve manipulation behavior through simulated interaction, building on the representations retained during compression.

We use \textbf{Proximal Policy Optimization (PPO)}~\cite{schulman2017proximal} with RLinf~\cite{zang2025rlinfvlaunifiedefficientframework} for parallel simulation rollouts and distributed policy updates. A sparse binary reward assigns 1 to successful task completion and 0 otherwise, providing a direct training signal without task-specific dense reward shaping. Rewards and log-probabilities are aggregated at the action-chunk level. PPO maximizes the clipped policy surrogate
\begin{equation}
\begin{split}
    J_{clip}(\theta)=\mathbb{E}_t\big[\min\big(&r_t(\theta)\hat{A}_t,\\
    &\operatorname{clip}(r_t(\theta),1-\delta,1+\delta)\hat{A}_t\big)\big],
\end{split}
\end{equation}
where $s_t$ is the policy's multimodal observation, $a_t$ is an action chunk, and $r_t(\theta)=\pi_\theta(a_t\mid s_t)/\pi_{\theta_{old}}(a_t\mid s_t)$ compares its probability under the current and rollout policies. The normalized generalized advantage estimate $\hat{A}_t$ guides the policy update, while $\delta$ controls the clipping range. Using task-completion feedback, the actor--critic loss trains the action expert and value head while keeping the pruned backbone frozen. The architecture remains fixed; paired evaluations before and after RL measure improvement over PAT only.

\section{Experiments}

\begin{table*}[!t]
    \centering
    \caption{\textbf{Robustness of Unpruned and Pruned $\pi_{0.5}$ on LIBERO-Plus~\cite{fei2025liberoplus}.}
    For each configuration and RL setting, we evaluate the same checkpoint on LIBERO and LIBERO-Plus. Retention targets, parameter counts, and pruning ratios follow \Cref{tab:main_results_int_hid}. Scores are percentages, with cross-suite results by perturbation and the aggregate under Overall. Parentheses show RL gains in percentage points. Bold highlights the best pruned score within each RL setting.}
    \label{tab:libero_plus}
    \footnotesize
    \setlength{\tabcolsep}{3pt}
    \begin{tabular*}{\textwidth}{@{\extracolsep{\fill}}llccccccccc@{}}
        \toprule
        \textbf{Model} & \textbf{Method} & $\boldsymbol{\rho_{hid}/\rho_{int}}$ & \textbf{Camera} & \textbf{Robot} & \textbf{Language} & \textbf{Light} & \textbf{Background} & \textbf{Noise} & \textbf{Layout} & \textbf{Overall} ($\boldsymbol{\Delta}$) \\
        \midrule
        $\pi_{0.5}$ & baseline & 100\% / 100\% & 71.04 & 74.77 & 85.82 & 95.80 & 94.89 & 86.51 & 85.84 & 83.98 \\
        \midrule
        \multirow{5}{*}{$\pi_{0.5}$} & \multirow{5}{*}{\textbf{PAT only}} & 90\% / 10\% & \textbf{60.60} & \textbf{54.65} & \textbf{68.77} & \textbf{92.03} & \textbf{87.27} & \textbf{76.89} & \textbf{78.43} & \textbf{72.68} \\
        & & 70\% / 10\% & 59.60 & 41.23 & 56.60 & 87.83 & 79.65 & 75.27 & 70.49 & 65.82 \\
        & & 50\% / 10\% & 56.66 & 32.58 & 52.37 & 82.14 & 71.47 & 66.83 & 62.95 & 59.35 \\
        & & 30\% / 10\% & 58.47 & 30.19 & 41.77 & 82.84 & 65.43 & 64.21 & 59.15 & 56.08 \\
        & & 10\% / 10\% & 59.47 & 24.00 & 31.36 & 76.71 & 48.98 & 64.15 & 34.43 & 47.46 \\
        \midrule
        $\pi_{0.5}$ & baseline+RL & 100\% / 100\% & 69.04 & 77.94 & 84.26 & 94.92 & 96.19 & 88.82 & 86.69 & 84.45\,\textcolor{darkgreen}{(+0.47)} \\
        \midrule
        \multirow{5}{*}{$\pi_{0.5}$} & \multirow{5}{*}{\textbf{PAT+RL}} & 90\% / 10\% & 65.92 & \textbf{62.00} & \textbf{76.06} & \textbf{95.27} & \textbf{91.82} & 82.01 & \textbf{83.28} & \textbf{78.20}\,\textcolor{darkgreen}{(+5.52)} \\
        & & 70\% / 10\% & 64.98 & 52.00 & 64.87 & 93.78 & 89.41 & \textbf{82.32} & 74.95 & 73.14\,\textcolor{darkgreen}{(+7.32)} \\
        & & 50\% / 10\% & 61.79 & 41.68 & 58.04 & 89.14 & 81.23 & 75.27 & 67.28 & 66.29\,\textcolor{darkgreen}{(+6.94)} \\
        & & 30\% / 10\% & 64.98 & 38.84 & 45.93 & 90.37 & 73.42 & 72.77 & 62.69 & 62.71\,\textcolor{darkgreen}{(+6.63)} \\
        & & 10\% / 10\% & \textbf{66.23} & 34.77 & 37.54 & 85.73 & 58.27 & 75.58 & 46.36 & 56.81\,\textcolor{darkgreen}{(+9.35)} \\
        \bottomrule
    \end{tabular*}
\end{table*}

\begin{table*}[!t]
    \centering
    \caption{\textbf{Real-Robot Task Results.} We evaluate unpruned and pruned $\pi_{0.5}$ on an AgileX dual-arm platform and report successes out of 20 trials per task. Target retention ratios and parameter reductions follow \Cref{tab:main_results_intermediate}. Avg. is the mean success rate across the six tasks described below.}
    \label{tab:real_robot}
    \footnotesize
    \setlength{\tabcolsep}{2pt}
    \begin{tabular*}{\textwidth}{@{\extracolsep{\fill}}llcccccccccc@{}}
        \toprule
        \multirow{2}{*}{\textbf{Model}} & \multirow{2}{*}{\textbf{Method}} & \multirow{2}{*}{$\boldsymbol{\rho_{hid}/\rho_{int}}$} & \multicolumn{2}{c}{\textbf{Pruning Ratio}} & \multirow{2}{*}{\shortstack{\textbf{Task 1}\\\textbf{Bottle}}} & \multirow{2}{*}{\shortstack{\textbf{Task 2}\\\textbf{Cubes}}} & \multirow{2}{*}{\shortstack{\textbf{Task 3}\\\textbf{Cup}}} & \multirow{2}{*}{\shortstack{\textbf{Task 4}\\\textbf{Block}}} & \multirow{2}{*}{\shortstack{\textbf{Task 5}\\\textbf{Sorting}}} & \multirow{2}{*}{\shortstack{\textbf{Task 6}\\\textbf{Backpack}}} & \multirow{2}{*}{\textbf{Avg. (\%)}} \\
        \cmidrule(lr){4-5}
        & & & \textbf{Model} & \textbf{LLM} & & & & & & & \\
        \midrule
        $\pi_{0.5}$ & baseline & 100\% / 100\% & 0\% & 0\% & 19/20 & 17/20 & 15/20 & 18/20 & 12/20 & 4/20 & 70.83 \\
        $\pi_{0.5}$ & PAT & 100\% / 20\% & 40\% & 58\% & 19/20 & 16/20 & 13/20 & 16/20 & 9/20 & 3/20 & 63.33 \\
        \bottomrule
    \end{tabular*}
    \par\smallskip
    \begin{tabularx}{\textwidth}{@{}l>{\raggedright\arraybackslash}X@{\hspace{12pt}}l>{\raggedright\arraybackslash}X@{}}
    \textbf{Task} & \textbf{Instruction} & \textbf{Task} & \textbf{Instruction} \\
    Task 1 & Pick up the bottle and place it upright. & Task 4 & Put the red block into the basket. \\
    Task 2 & Place the blue cube to the right of the yellow cube. & Task 5 & Put the screws in the blue basket and the blocks in the red basket. \\
    Task 3 & Pick up the cup and place it on the plate. & Task 6 & Put the object in the backpack. \\
    \end{tabularx}
\end{table*}

\begin{figure*}[!t]
    \centering
    \includegraphics[width=\linewidth]{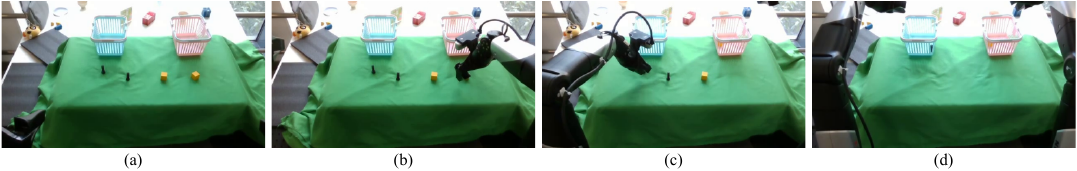}
    \caption{\textbf{Real-Robot Sorting Task on the AgileX Dual-Arm Platform.}
    Four frames from one recorded execution of Task 5: (a) initial arrangement, (b) block manipulation, (c) screw manipulation, and (d) the final recorded scene. Quantitative results are reported in \Cref{tab:real_robot}.}
    \label{fig:real_robot_sequence}
\end{figure*}

\paragraph{Tasks and Training Protocol}
We train one policy per configuration across all 40 tasks in LIBERO's Spatial, Object, Goal, and Long (LIBERO-10) suites~\cite{NEURIPS2023_8c3c6668}, covering spatial, object, goal, and long-horizon variation. Using OpenPI's PyTorch $\pi_{0.5}$ implementation~\cite{physicalintelligence2025openpi} with a Gemma 2B backbone, we vary retention targets for the hidden and intermediate dimensions and report suite success rates and their mean.

\paragraph{Baselines}
We compare with unpruned $\pi_{0.5}$ before and after RL and published results for NORA~\cite{hung2025norasmallopensourcedgeneralist}, NORA-1.5~\cite{hung2025nora15}, SmolVLA~\cite{shukor2025smolvlavisionlanguageactionmodelaffordable}, GraspVLA~\cite{deng2025graspvla}, and EcoVLA~\cite{huang2026ecovla}. SPAR's parameter-pruning ratios are relative to the unpruned full model (3.37B) and LLM (2.34B), with pruning percentages rounded to integers. EcoVLA's labels denote its reported channel/head pruning settings.

\paragraph{Implementation Details}
SFT uses a batch size of 256 for 30,000 steps: 6,000 for PAT mask learning and 24,000 for post-pruning fine-tuning. Training and RL use NVIDIA A800 GPUs. We deploy on NVIDIA Jetson AGX Orin and measure median full-policy inference latency using TensorRT.

\paragraph{PPO Configuration}
PPO uses eight NVIDIA A800 GPUs on one node, with 128 training and 80 evaluation environments across all four suites. The actor global batch size is 2,048, with a per-rank micro-batch size of 128 and two gradient-accumulation steps. The runner allows up to 200 RL epochs.

\subsection{Main Results}

Pruning only the intermediate dimension preserves high task success (\Cref{tab:main_results_intermediate}). A 10\% intermediate-dimension retention target reduces the LLM from 2.34B to 0.82B with 96.45\% average success, only 0.35 percentage points below the baseline's 96.80\%. This suggests substantial FFN intermediate-dimension redundancy on these tasks, motivating $\rho_{int}=0.1$ for subsequent experiments.

Additional hidden-dimension pruning causes larger losses (\Cref{tab:main_results_int_hid}): reducing $\rho_{hid}$ from 0.9 to 0.1 lowers average success without RL from 95.90\% to 84.55\%, most sharply on Long. RL substantially mitigates these losses, improving all five pruned averages by 1.90--8.90 percentage points versus 0.90 for the baseline. Notably, at $\rho_{hid}=0.7$ and $\rho_{int}=0.1$, RL achieves 97.55\%, just 0.15 points below the unpruned RL baseline's 97.70\%, with approximately 52\% fewer total and 75\% fewer LLM parameters. Even the smallest policy (1.12B total / 0.09B LLM) improves from 84.55\% to 93.45\%, recovering much of the loss while remaining 4.25 points below that baseline.

\subsection{Robustness on LIBERO-Plus}

LIBERO-Plus~\cite{fei2025liberoplus} tests changes in camera viewpoints, robot initial states, language instructions, lighting, background textures, sensor noise, and object layouts. \Cref{tab:libero_plus} reports cross-suite scores by perturbation category and the aggregate score under Overall.

RL substantially improves all five pruned overall scores by 5.52--9.35 percentage points, versus 0.47 for the baseline. The smallest policy improves from 47.46\% to 56.81\% but remains 27.64 points below the unpruned RL baseline, particularly under robot-initial-state and language perturbations. Even at $\rho_{hid}=0.9$ and $\rho_{int}=0.1$, the recovered policy scores 78.20\% versus the baseline's 84.45\%. This 6.25-point gap, despite nearly matched standard LIBERO performance, indicates greater sensitivity to perturbations after compression. Robustness-critical applications may therefore require less aggressive pruning or further RL adaptation with broader perturbation exposure.

\subsection{Real-Robot Evaluation}
On an AgileX dual-arm platform, we evaluate six tasks with 20 trials each (\Cref{tab:real_robot}); \Cref{fig:real_robot_sequence} illustrates Task 5. Pruning only the intermediate dimension with $\rho_{int}=0.2$ yields 2.02B total / 0.99B LLM parameters, approximately 40\% / 58\% fewer than the baseline. The compact policy retains much of the baseline's manipulation performance: 76/120 successes (63.33\%) versus 85/120 (70.83\%), a 7.50-percentage-point decrease. It matches the baseline on Task 1 and records successes on all six tasks. These results show that intermediate-dimension pruning preserves practical manipulation capability on physical hardware despite substantial parameter reduction.

\Needspace{4\baselineskip}
\subsection{Inference Latency}

\begin{table}[!ht]
    \centering
    \caption{\textbf{Inference Latency and GPU Memory of $\pi_{0.5}$.}
    We report median full-policy latency and GPU memory using TensorRT on NVIDIA Jetson AGX Orin. Memory is in GiB, speedup is the unpruned-to-pruned latency ratio, and targets follow \Cref{tab:main_results_intermediate}. Best values are bold.}
    \label{tab:main_results_speed}
    \footnotesize
    \setlength{\tabcolsep}{3pt}
    \begin{tabular}{@{}lcccc@{}}
        \toprule
        \textbf{Model} & $\boldsymbol{\rho_{hid}/\rho_{int}}$ & \shortstack{\textbf{Latency}\\ \textbf{(ms, $\downarrow$)}} & \textbf{Speedup ($\uparrow$)} & \shortstack{\textbf{GPU Memory}\\ \textbf{(GiB, $\downarrow$)}} \\
        \midrule
        \rowcolor{gray!10} $\pi_{0.5}$ & 100\% / 100\% & 295.04 & 1.00 & 7.098 \\
        \midrule
        \multirow{6}{*}{$\pi_{0.5}$} & 100\% / 90\% & 282.32 & 1.05 & 6.713 \\
        & 100\% / 70\% & 252.36 & 1.17 & 6.039 \\
        & 100\% / 50\% & 222.65 & 1.33 & 5.395 \\
        & 100\% / 30\% & 194.20 & 1.52 & 4.752 \\
        & 100\% / 20\% & 178.93 & 1.65 & 4.400 \\
        & 100\% / 10\% & 170.64 & 1.73 & 4.086 \\
        \midrule
        \multirow{5}{*}{$\pi_{0.5}$} & 90\% / 10\% & 165.52 & 1.78 & 3.908 \\
        & 70\% / 10\% & 162.76 & 1.81 & 3.596 \\
        & 50\% / 10\% & 156.10 & 1.89 & 3.281 \\
        & 30\% / 10\% & 151.57 & 1.95 & 2.973 \\
        & 10\% / 10\% & \textbf{148.71} & \textbf{1.98} & \textbf{2.627} \\
        \bottomrule
    \end{tabular}
\end{table}

\Cref{tab:main_results_speed} reports full-policy latency and memory on NVIDIA Jetson AGX Orin. At 10\% intermediate retention, latency falls from 295.04\,ms to 170.64\,ms (1.73$\times$); adding hidden-dimension pruning at the same retention reaches 148.71\,ms, a $1.98\times$ speedup and 49.60\% latency reduction. GPU memory decreases from 7.098 to 2.627\,GiB, approximately 63\% lower than the baseline.

\Needspace{4\baselineskip}
\subsection{Ablation Study}

\begin{table}[!ht]
    \centering
    \caption{\textbf{Pruning-Loss Ablation for $\pi_{0.5}$, Evaluated on LIBERO.}
    We remove one loss at a time from the full objective, targeting $\rho_{hid}=100\%$ and $\rho_{int}=10\%$ without RL. Scores are percentages, Long denotes LIBERO-10, and Avg. is the four-suite mean. Bold identifies the best value in each column.}
    \label{tab:ablation_soft_mask}
    \footnotesize
    \setlength{\tabcolsep}{4pt}
    \begin{tabular}{@{}lccccc@{}}
        \toprule
        \textbf{Method} & \textbf{Spatial} & \textbf{Object} & \textbf{Goal} & \textbf{Long} & \textbf{Avg.} \\
        \midrule
        PAT w/o $\mathcal{L}_{count}$ & \textbf{97.60} & 98.60 & \textbf{97.40} & 89.20 & 95.70 \\
        PAT w/o $\mathcal{L}_{entropy}$ & 97.40 & 98.40 & 96.60 & 92.00 & 96.10 \\
        PAT w/o $\mathcal{L}_{disc}$ & \textbf{97.60} & \textbf{98.80} & 97.20 & 90.60 & 96.05 \\
        \midrule
        PAT (full objective) & 97.40 & 97.60 & 96.80 & \textbf{94.00} & \textbf{96.45} \\
        \bottomrule
    \end{tabular}
\end{table}

\Cref{tab:ablation_soft_mask} shows that the full objective in \Cref{tab:main_results_intermediate} achieves the highest average success rate. Removing the count, entropy, or discreteness term lowers the average by 0.75, 0.35, or 0.40 percentage points, respectively. The full objective also achieves the highest Long-suite score, supporting the combined use of budget and binarization penalties in this configuration.

\section{Limitations and Future Work}

Although SPAR approaches or slightly exceeds the unpruned RL baseline on LIBERO under some aggressive compression settings, a robustness gap remains on LIBERO-Plus. Evaluation is also limited to one backbone and one intermediate-dimension pruning configuration across six real-robot tasks, motivating broader backbone studies, joint hidden-dimension and intermediate-dimension pruning on robots, and energy measurements across deployment platforms.

To further improve recovery, we envision using SimFoundry~\cite{ranawaka2026simfoundry} to reconstruct real-world AgileX task scenes as simulation environments. TiPToP~\cite{shen2026tiptopmodularopenvocabularyrobot} and ASPIRE~\cite{lu2026aspire} could then generate trajectories through planning and program execution, with MimicGen~\cite{mandlekar2023mimicgen} expanding their coverage for supervised adaptation of compact policies. Building on this initialization, the policies would undergo online reinforcement learning with PPO through direct interaction in the real-to-sim environments reconstructed by SimFoundry.

\section{Conclusion}
SPAR jointly prunes the hidden dimension and intermediate dimension with separate retention targets, combining pruning-aware adaptation with PPO-based recovery. Intermediate-only pruning preserves high success, while RL partially restores hidden-dimension losses. With a 0.09B LLM and 1.12B total parameters, the compact policy achieves a $1.98\times$ inference speedup, while RL raises its LIBERO success from 84.55\% to 93.45\%. On AgileX, 20\% intermediate retention leaves a 7.50-point gap to the baseline. These results support combining dimension-aware compression and RL to improve the performance--size trade-off of VLA models.

\bibliographystyle{IEEEtran}
\bibliography{SPAR}
\end{document}